\documentclass[10pt,twocolumn,letterpaper]{article}

\usepackage{iccv}
\usepackage{times}
\usepackage{epsfig}
\usepackage{graphicx}
\usepackage{amsmath}
\usepackage{amssymb}
\usepackage{tabu}
\usepackage{url}
\usepackage{booktabs}
\usepackage{graphicx}
\usepackage{glossaries}

\graphicspath{{figures/}}

\newacronym{cnn}{CNN}{Convolutional Neural Networks}

\usepackage[pagebackref=true,breaklinks=true,letterpaper=true,colorlinks,bookmarks=false]{hyperref}

\iccvfinalcopy 

\begin{document}

\title{Skip-Clip: Self-Supervised Spatiotemporal Representation Learning  by Future Clip Order Ranking}

\author{
	Alaaeldin El-Nouby$^{1,4}$\thanks{Work was performed during an internship 
	with Apple Inc.} \quad Shuangfei Zhai $^{2}$ \quad    Graham 
	W.~Taylor$^{1,3,4}$ \quad Joshua M. Susskind$^{2}$ \\\vspace*{1mm}
	{$^{1}$ University of Guelph \quad
		$^{2}$ Apple Inc. \quad }
		\\	
		{$^{3}$ Canadian Institute for Advanced Research \quad
		$^{4}$ Vector Institute for Artificial Intelligence
		}
}

\maketitle

\begin{abstract}

Deep neural networks require collecting and annotating large amounts of data to train successfully. In
order to alleviate the annotation bottleneck, we propose a novel self-supervised
representation learning approach for spatiotemporal features extracted from
videos. We introduce Skip-Clip, a method that utilizes temporal coherence in
videos, by training a deep model for future clip order ranking conditioned on
a context clip as a surrogate objective for video future prediction. We show
that features learned using our method are generalizable and transfer strongly
to downstream tasks. For action recognition on the UCF101 dataset, we obtain 51.8\%
improvement over random initialization and outperform models initialized using
inflated ImageNet parameters. Skip-Clip also achieves results competitive with state-of-the-art self-supervision methods.
\end{abstract}

\section{Introduction}

The performance of deep \gls{cnn}s relies heavily on the availability
of human annotations to power supervised learning. While visual data like
images and videos are abundant, providing semantic labels for such large
amounts of data can be very expensive and time-consuming. There is thus a need for
methods that can utilize huge amounts of available unlabeled data. This will
improve the scalability of deep learning methods and make them more accessible
to new domains that suffer from high annotation costs. Researchers have
studied different approaches to enable learning with less dependence on
labels. 

There is a whole class of algorithms that follow the unsupervised learning
paradigm. One interesting example of such approaches is self-supervised
learning  \cite{context_video_wang, context_pred, context_encoder, shuffle,
arrow, color_1}. For self-supervision methods, a pretext task is designed to
exploit structure in data. Pseudo-labels are generated automatically from the
data structure and a deep model can be trained to minimize the loss with
respect to the generated labels, using standard supervised learning methods.

\begin{figure}[t]
    \centering
    \includegraphics[scale=1.45]{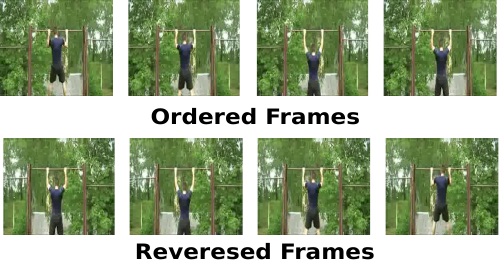}
    \caption{A short 4 frame clip of a person performing a pull-up action.
    \textit{Left} are the frames in the correct order from left to right. On
    the \textit{Right} the same frames but in the reverse order. It is clear
    that both set of frames are temporally plausible when observed out of
    context.}
     \label{order_matters}
\end{figure}
 
The study of the video domain in computer vision is of great importance. It
powers computer-human interaction applications as well as enables
implicit learning of laws of motion and physics. However, providing annotations
for videos is particularly
difficult due to its temporal dimension. Additionally, models used for 
video-related tasks like action recognition are usually heavily parameterized 
3D \gls{cnn}s \cite{Hara_2018} that enable simultaneous modelling of spatial
and temporal dimensions. Training high capacity models requires a huge amount of
data to prevent over-fitting. Until recently, training of 3D \gls{cnn}s have not been very successful due to the lack of large scale datasets. Considering
the high cost of annotations and the strong need for huge amounts of data,
learning from videos can gain significantly from a strong self-supervision
method. Videos are usually temporally coherent and there is a high correlation
in the spatial information across nearby frames. This property can be a source
of self-supervision signals as shown by \cite{arrow, shuffle, cubic_puzzle,
track_color}. Some efforts utilize the temporal order signal by
training a model to either verify the order of frames \cite{arrow, shuffle} or to sort
a shuffled set of frames \cite{OPN}. Such tasks can provide a strong
supervisory signal, however,
setting up the pretext task this way can lead to many noisy samples. For
example in Figure~\ref{order_matters} we show two sets of ordered and
reversed frames of a person doing a pull-up action, both sets are temporally
plausible since \emph{pull-up} is a cyclic action. Training a model for an
ambiguous
label can hinder the strength and the generality of the representation. The
reason this task can be ambiguous is that we train a model to sort frames out
of context. There is no information about how the current state of the scene
came to be.

In this work, we propose a method that alleviates the shortcomings of sorting and future frame prediction approaches by combining both ideas in a
single method. From a given video, we sparsely sample a set of frames to be
sorted as well as a set of contiguous surrounding frames as a context. Instead
of training a model to predict the correct ordering out of a disjoint set
of different possible orderings, the model is trained to predict the correct
relative position of every frame given the context using a ranking objective.
Ranking of future frames encourages the model to learn about scene dynamics 
and  tracking without requiring expensive full frame prediction.
Training the model to rank future frames given a context is a softer, more
controlled instantiation of future prediction in latent space.

We summarize our contribution as follows:
\begin{itemize}
\item We propose a novel pretext task for spatiotemporal representation
learning. Our method combines the two ideas of predicting future frames and
exploiting temporal coherence, by learning to sort frames in a simple and
efficient framework.
\item In order to demonstrate the quality of the learned representations, we
provide strong results for the downstream task of action recognition using the UCF101 dataset. \end{itemize}{}

\section{Related Work}

In the language domain, self-supervised methods for learning word or
sentence representations have been dominant. One of the most successful approaches
learns an encoding per word that maximizes the probability of its
surrounding words. Representations learned using this objective encode strong
semantics about language. \cite{mikolov2013efficient} proposed the Skip-gram
model in which a context word is encoded in a way that allows predicting surrounding words. The same idea has
been extended to sentences \cite{kiros2015skipthought}. Our method can be seen as an instantiation of this idea adapted to the video domain, hence
we name our method Skip-Clip. In this work, we focus on predicting future clips only, though the method can easily be extended to consider both future and past clips.

One of the related methods to our approach is Contrastive Predictive Coding 
(CPC) \cite{Oord2018RepresentationLW}. It studies unsupervised representation
learning by predicting the future in the latent space by using a contrastive
loss between positive future samples and negative samples sampled from a
random different sequence. While this method has shown promising results on
different domains like speech, images, and text, applying this method to
videos reduces to a trivial task since there is a strong visual similarity
between the positive samples and the context. In our method, we also predict
the future in a latent space. However, our objective is a hinge rank loss in
which we rank positive future samples with respect to their temporal distance
from the context. This task is significantly harder since we train the
model to learn a latent space that represents differences in very similar
frames in order to be successful at the ranking task. Accordingly, an encoder
that learns such a latent space capturing the subtleties needed for fine-grained frame ranking will be useful for
downstream tasks.

\section{Methodology}

\begin{figure*}[t]
    	\centering
    	\includegraphics[scale=0.5]{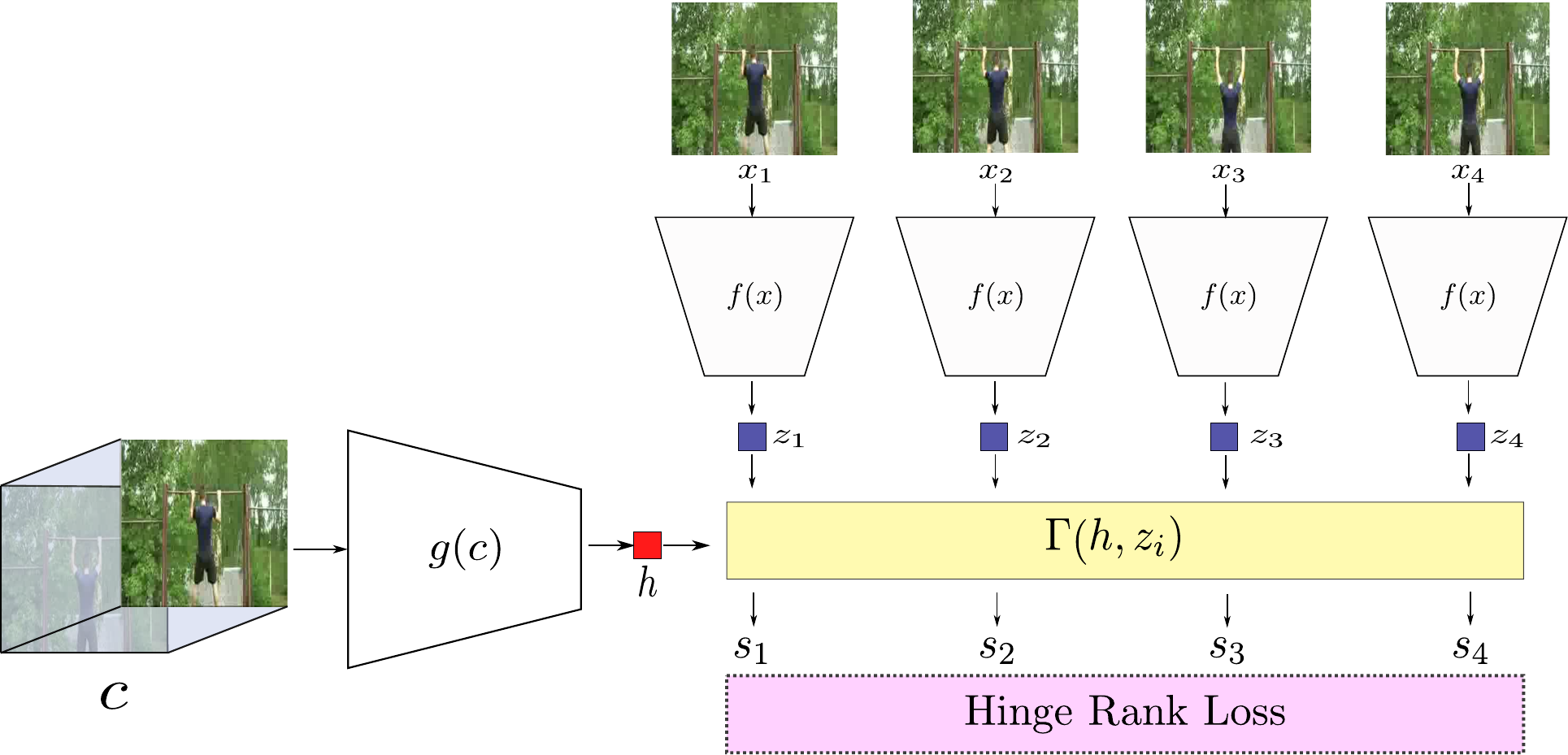}
    	\caption{A context clip $c$ 
    	is sampled as well as $M$ clips from the following frames $\{x_{1}, \cdots, 
    	x_{M}\}$. Latent representations $h$ and $z_{i}$ are extracted using 
    	encoding functions $g(c)$ and $f(x)$ for the context and target clips 
    	respectively. A scoring function $\Gamma$ measures the ranking score 
    	between the target representations relative to the context. Finally, 
    	the scores are used in hinge rank losses (equation~\ref{hinge_rank} and \ref{contrastive}).}
    	\label{skipclip}
    	\vspace{-0.3cm}
 \end{figure*}

We propose a simple and powerful framework for self-
supervised spatiotemporal representation learning by combining two core ideas:
predicting the future based on context, and temporal consistency by ordering
clips. An overview of our method is illustrated in Figure~\ref{skipclip}.

\subsection{Skip-Clip}

Given a video consisting of a set of $N$ frames $V = \{v_{1}, v_{2}, \cdots,
v_{N} \}$, we randomly seek to time step $t$ in the video \footnote{Random 
seeking for $t$ has to be a certain distance from the end of the video to 
enable sampling context and target clips}. We densely sample
$K$ context frames $c = \{v_{t}, v_{t+1}, \cdots ,v_{t+K}\}$. We sparsely
sample $M$ target clips $T=\{x_{1}, x_{2}, \cdots, x_{M}\}$ using fixed
sampling rate $r$ such that all target clips are subsequent to the context
frames. The context clip and the first target clip are $r$ frames apart.
Each target clip contains $d$  contiguous frames (target clip length
is $d$). Additionally, we
sample another $M$ clips from different videos to create a set of clips $Q =
\{\overline{x}_{1}, \overline{x}_{2}, \cdots, \overline{x}_{M}\}$ that can be
used as negative samples.

We define encoding functions for the context and the target clips to obtain a
compressed representation in the latent space. For the context, we have $h =
g(c)$ where $h$ is the latent representation of the context clip. Similarly
each clip in the target set $T$ and negative set $Q$ is encoded as
$z_{i}=f(x_{i})$ and $\overline{z}_{i}=f(\overline{x}_{i})$ respectively. Our
goal is to train the encoding functions $g(c)$ and $f(x)$ to enable a
successful ranking of target clips conditioned on the context clip.
We then need to define a scoring function $s = \Gamma(h,
z_{i})$ that describes the  relationship between the context and target
encodings. The scores for high ranking targets should be
higher than low ranking ones.  We implement this objective using a hinge rank
loss:

 \begin{equation}
 \label{hinge_rank}
     L_{\text{rank}} = \sum_{i=0}^{M-1} \sum_{j=i+1}^{M} \max(0, - \Gamma(h, z_{i}) + 
     \Gamma(h, z_{j}) + 
     \delta_{rank}).
 \end{equation}{}

\noindent where $\delta_{rank}$ is the margin. This objective is challenging
since the target clips share high spatial similarity with the context. For
successful ranking, the  latent representations need to encode information
about the various motions of objects and scene and be aware of physical concepts like
velocity to determine the correct position of a target clip relative to the
context clip.

Additionally, scores for target clips should be higher than 
negative clips given the context encoding. Therefore we introduce a contrastive
loss:

  \begin{equation}
  \label{contrastive}
     L_{\text{contrastive}} = \sum_{i=0}^{M} \max(0, -\Gamma(h, z_{i}) + \mathbb{E}_{\overline{z} \in  
     Q}[\Gamma(h, \overline{z})] + \delta_{neg}).
 \end{equation}{}

This is important because the ranking objective
can choose to focus mainly on temporal cues since it is more strongly correlated
with motion. An objective that forces the encodings to discriminate between
two spatially dissimilar clips is required to also pick up on spatial subtleties. If we train our model using only 
the contrastive objective in equation~\ref{contrastive}, our model will be equivalent to the 
the CPC model \cite{Oord2018RepresentationLW}.

Finally, to enrich the spatial signal, we add
a rotation prediction auxiliary task. In experiments presented below, we found that
by training $f(x)$ solely for the ranking and contrastive objectives, it achieved
strong ranking performance that did not translate to better generalization to
the downstream task of action recognition, indicating that with the large capacity
of $g(c)$ and $f(x)$, the model was able to learn trivial solutions to the ranking
task. Adding an auxiliary objective helped restrict the representations learned by 
$f(x)$ such that it retains the spatial information. Training the two encoders for ranking with the presence of these restrictions led to better correlation between the ranking task and generalization to the downstream task.

\section{Implementation Details}

\subsection{Self-supervision}

For the context clip, we sample $K=16$ contiguous frames. $g(c)$ is implemented 
using a 3D \gls{cnn}, specifically, a 3D ResNet-18  \cite{Hara_2018} up to the 
last convolutional layer. As for the target clips, we implement the special 
case where $d=1$ such that each target clip is a single frame. Empirically, we found that using longer target clips gives the model more room to exploit trivial solutions for the ranking task that does not generalize well. Ranking is performed on $M=8$ target clips. $f(x_{i})$ is implemented using a 2D ResNet-18 
\cite{resnet}. Both encoders are trained from scratch (random initialization). Both the context and target clip encodings $h$ and $z_{i}$ are 
tensors of dimensionality $\mathcal{R}^{C\times H\times W}$ where number of 
channels $C=512$ and
the spatial dimensions $H \times W = 7\times 7$. We found that keeping the 
spatial dimension is more informative when used with the scoring 
function compared to having the encodings in vector form. For the scoring 
function $\Gamma(h, z_{i})$, given two $\mathcal{R}^{C\times H\times W}$ 
tensors, we compute 
the average cosine similarity across every two aligned spatial cells in the 
$H\times W$ grid: 

\begin{equation}
\label{cosine}
\Gamma(h, z_{i}) = \frac{1}{H * W}\sum_{m=0}^{H} \sum_{n=0}^{W} \frac { h^{m, 
n} 
\cdot  z^{m, n}_{i}}{|| h^{m, n}|| \cdot || z^{m, n}_{i}||}
\end{equation}

\noindent where, $h^{m,n}$ indicates the 
vector in the $m^{th}$ row and $n^{th}$ column with dimension  $\mathcal{R}^C$.
For the hinge rank losses in equations~\ref{hinge_rank} and \ref{contrastive}.

The rotation auxiliary objective  follows
\cite{rotnet} predicting a rotation angle out of the fixed set  [0$^{\circ}$,
90$^{\circ}$, 180$^{\circ}$, 270$^{\circ}$]. All loss terms are summed without
any weighting. More details about the hyper-parameters used for pre-training can be found in the supplementary materials.

\subsection{Action Recognition fine tuning}

To demonstrate the quality of the representations learned using the pretext
task, we fine tune the context clip encoder $g(c)$ for the downstream task of
action recognition. We add a global average pooling layer followed by a
fully-connected layer with softmax outputs to the encoder network. We use split 1 of
the UCF101 dataset for training and testing. $g(c)$ is a 3D \gls{cnn}. We use
an input of 16 frames with spatial dimensions of 112$\times$112 after applying
random cropping. We fine tune for 300 epochs. We use the Adam optimizer with
weight decay of $1e^{-2}$ and learning rate of $5e^{-4}$ that is multiplied by
0.5 every 15 epochs up to the 60$^{th}$ epoch.  For testing, we use center cropping and apply a
sliding window approach by averaging the softmax outputs over all the 16 frame
non-overlapping windows in a given test video.

\section{Experiments and Results}

We evaluate our model Skip-Clip for the downstream task of action
recognition using the UCF101 dataset. We compare our results to different
initialization baselines as shown in Table~\ref{baselines}. First, we notice
that using Skip-Clip is 
clearly superior to training from scratch. Our model achieves accuracy that is
51.8\% higher than a model trained from random initialization of the weights.
Secondly, our model outperforms a model initialized with inflated ImageNet
\cite{DBLP:journals/corr/CarreiraZ17} weights while being completely
unsupervised during pre-training.

\begin{table}
	\centering
	\begin{tabular}{l c c c}
		\toprule
		\textbf{Method} & \textbf{Accuracy}  \\
		\midrule
		Random Initialization \cite{Hara_2018} &  42.4 \\
		ImageNet Inflation \cite{cubic_puzzle}  & 60.3 \\
		Skip-Clip	 & \textbf{64.4}  \\
		\bottomrule
	\end{tabular}
	\caption{ Top-1 accuracy comparison to standard initialization baselines 
	performance for action recognition task on UCF-101 dataset. }
	\label{baselines}
\end{table}{}

We study the contribution of multiple components in Table~\ref{ablation}. We observe that a basic Skip-Clip model without the auxiliary rotation objective can lead to relatively weaker performance. This is because the basic model is susceptible to learning trivial solutions to the ranking task, which does not necessarily transfer well to other semantic tasks like action recognition. By adding the rotation auxiliary objective to the target encoder $f(x)$, we see a significant improvement of 3.6\%. Furthermore, by adding the contrastive loss as described in equation \ref{contrastive}, we obtain a strong performance of 64.4\%.

\begin{table}
    \small
    \centering
    \begin{tabular}{l c c c}
        \toprule
         \textbf{Model} & \textbf{UCF101} \\
         \midrule
		 Skip-Clip  & 59.5 \\
  		 Skip-Clip + rotation  & 63.1  \\
		 Skip-Clip + rotation + negative sampling & \textbf{64.4} \\

         \bottomrule
    \end{tabular}
    \caption{Ablation Study comparing the base model to models with additional auxiliary objectives.}
    \label{ablation}
\end{table}{}

We compare Skip-Clip to other self-supervision methods as shown in
Table.~\ref{results_sota}. Skip-Clip significantly outperforms all 2D \gls{cnn}
based methods \cite{shuffle, arrow, OPN}. Additionally, it achieves competitive performance compared to the methods that used UCF101 for pre-training
using the self-supervised objective, outperforming
\cite{video_gan, wang2019selfsupervised} and being on par with \cite{clip_order}. The strong performance of \cite{clip_order} suggests that there is potential in using target clips larger than a single frame in future work.
As for the methods
that used the large scale Kinetics dataset for pre-training, Skip-Clip
outperforms \cite{jing2018selfsupervised} despite using a significantly
smaller dataset for pre-training. Finally, we fall behind only the 3D Cubic
Puzzles \cite{cubic_puzzle} approach; however, this method takes
advantage of pre-training using Kinetics, which we will explore in future work.

\begin{table}
    \footnotesize
    \centering
    \begin{tabular}{l c c c c}
        \toprule
         \textbf{Method} & \textbf{Backbone} & \textbf{Source}& \textbf{UCF101} & \\
         \midrule
		 Shuffle and Learn \cite{shuffle} & AlexNet & UCF101 & 50.9 \\
         Arrow of time \cite{arrow} & AlexNet & UCF101 & 55.3\\
          OPN \cite{OPN} & AlexNet  & UCF101 & 56.3 \\
         \midrule
         VideoGAN \cite{video_gan} & C3D & UCF101  & 52.1 \\ 
         Motion \& Appearance \cite{wang2019selfsupervised} & C3D & UCF101 & 58.8 \\
         Motion \& Appearance \cite{wang2019selfsupervised} & C3D & Kinetics & 61.2\\
         3DRotNet \cite{jing2018selfsupervised} & 3D ResNet-18 & Kinetics &
         62.9 \\
          Video Clip Ordering \cite{clip_order} & R3D & UCF101 & 64.9 \\
         3DCubicPuzzles \cite{cubic_puzzle} & 3D ResNet-18 &Kinetics & \textbf{65.8} \\
         \midrule
         Skip-Clip	 & 3D ResNet-18 & UCF101 & 64.4  \\
         \\
         \bottomrule
    \end{tabular}
    \caption{Top-1 Accuracy performance for action recognition task on UCF-101 dataset. Different  backbones used for by the methods can account for some of the performance difference. }
    \label{results_sota}
\end{table}{}

\section{Conclusion and Future Work}

We have presented a method, Skip-Clip, for self-supervised spatiotemporal representation
learning. We combine the strengths of two popular ideas, future frame
prediction and temporal sorting of video frames. We demonstrate the strengths
and the generalization of the representations learned using our method by
finetuning the encoders for the downstream task of action recognition. We
show  that our method is competitive compared to other self-supervised
approaches. Additionally, our method outperforms the strong inflated ImageNet
baseline and beats some models that leverage the much larger-scale Kinetics
dataset. These results demonstrate the value of our proposition to pose frame
sorting as a predictive task that relies on a context.

{\small
\bibliographystyle{ieee}
\bibliography{egbib}
}

\clearpage
\twocolumn[{%
 \centering
 \bf
 \LARGE Appendix\\[1.5em]
}]
\appendix
\section{Qualitative Analysis}

\begin{figure*}[!h]
\centering
   \begin{tabular}{ccccc}
      \includegraphics[scale=0.35]{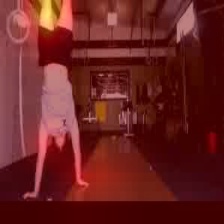} &
      \includegraphics[scale=0.35]{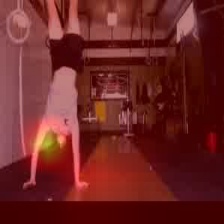} &
      \includegraphics[scale=0.35]{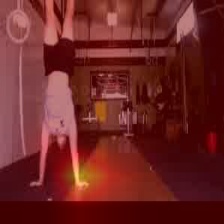} &
      \includegraphics[scale=0.35]{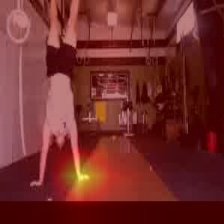} \\
       \includegraphics[scale=0.35]{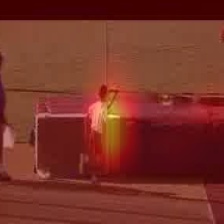} &
      \includegraphics[scale=0.35]{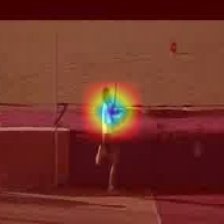} &
      \includegraphics[scale=0.35]{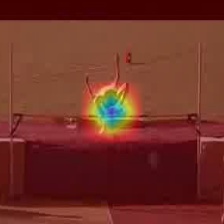} &
      \includegraphics[scale=0.35]{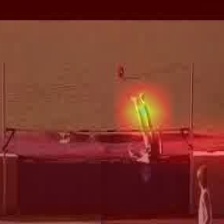} \\
       \includegraphics[scale=0.35]{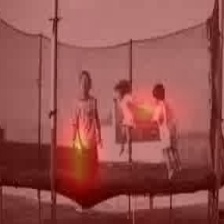} &
      \includegraphics[scale=0.35]{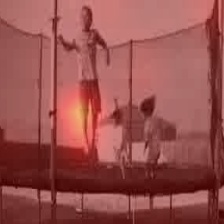} &
      \includegraphics[scale=0.35]{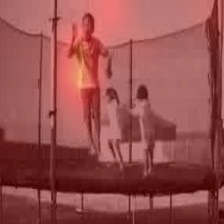} &
      \includegraphics[scale=0.35]{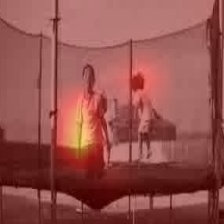} \\
       \includegraphics[scale=0.35]{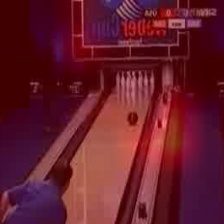} &
      \includegraphics[scale=0.35]{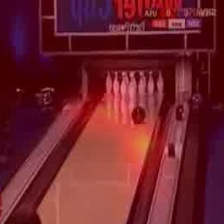} &
      \includegraphics[scale=0.35]{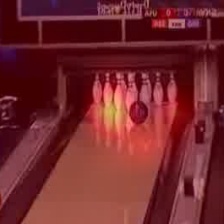} &
      \includegraphics[scale=0.35]{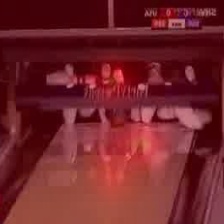} \\
   \end{tabular}
   \caption{Visualizations of the cosine similarity per aligned cell between context and target representations that are  part of computing the scores $\Gamma(h, z)$. We can observe that the highlighted regions correspond to salient motions in the frames.}
   \label{heatmaps}
\end{figure*}

To understand which regions of each frame contribute more to  frame ranking relative to the context, we visualize the distribution of the cosine similarity scores between $g(c)$ and $f(x)$ feature maps across different spatial locations in Figure \ref{heatmaps}. From these samples, it is clear that higher scores correlate with more salient regions in the frame in terms of object/person motion. We hypothesize that these regions are good cues for action recognition.

\section{Pre-training Implementation Details}

We used videos from split 1 of the training set of UCF101 \cite{ucf101} for training on
the pretext task. The input frames are resized to 224$\times$224 pixels. We
used the same hyperparameters to train all the encoder parameters. We
trained for 600 epochs using a batch size of 128. We used the Adam
optimizer \cite{DBLP:journals/corr/KingmaB14} with a learning rate of
$3e^{-4}$ that is multiplied by 0.1 every 200 epochs and weight decay of
$1e^{-7}$. For data augmentation, we randomly reversed the video frames
before sampling the context to double the motion patterns explored by the
model. Additionally, we applied random horizontal flipping and random cropping
consistent for the context and target clips. The target sampling rate was
set to $r=4$.

\end{document}